\documentclass[lettersize,journal]{IEEEtran}
\usepackage{amsmath,amsfonts}
\usepackage{algorithmic}
\usepackage{algorithm}
\usepackage{array}
\usepackage[caption=false,font=normalsize,labelfont=sf,textfont=sf]{subfig}
\usepackage{textcomp}
\usepackage{stfloats}
\usepackage{url}
\usepackage{verbatim}
\usepackage{graphicx}
\usepackage{cite}
\usepackage{makecell}
\usepackage{svg}
\usepackage{multirow}
\usepackage[colorlinks,
            linkcolor=red,
            anchorcolor=green,
            citecolor=blue
            ]{hyperref}
\usepackage{booktabs}
\usepackage{multirow}

\begin{document}

\title{Generalizable Deepfake Detection Based on Forgery-aware Layer Masking and Multi-artifact Subspace Decomposition}

\author{Xiang Zhang, Wenliang Weng, Daoyong Fu, Beijing Chen, Ziqiang Li, Ziwen He, Zhangjie Fu
        % <-this % stops a space    
\thanks{This work was supported in part by the National Natural Science Foundation of China under Grant 62202234; in part by the China Postdoctoral Science Foundation under Grant 2023M741778.

Xiang Zhang, Wenliang Weng, Daoyong Fu, Beijing Chen, Ziqiang Li, Ziwen He, and Zhangjie Fu are with the Engineering Research Center of Digital Forensics, Ministry of Education, Nanjing University of Information Science and Technology, Nanjing, 210044, China (e-mail: zhangxiang@nuist.edu.cn; 202412200722@nuist.edu.cn; fudymo@hotmail.com; nbutimage@126.com; iceli@mail.ustc.edu.cn; ziwen.he@nuist.edu.cn; fzj@nuist.edu.cn).}}

% The paper headers
\markboth{Journal of \LaTeX\ Class Files,~Vol.~14, No.~8, August~2021}%
{Shell \MakeLowercase{\textit{et al.}}: A Sample Article Using IEEEtran.cls for IEEE Journals}

% Remember, if you use this you must call \IEEEpubidadjcol in the second
% column for its text to clear the IEEEpubid mark.

\maketitle

\begin{abstract}
Deepfake detection remains highly challenging, particularly in cross-dataset scenarios and complex real-world settings. This challenge mainly arises because artifact patterns vary substantially across different forgery methods, whereas adapting pretrained models to such artifacts often overemphasizes forgery-specific cues and disturbs semantic representations, thereby weakening generalization. Existing approaches typically rely on full-parameter fine-tuning or auxiliary supervision to improve discrimination. However, they often struggle to model diverse forgery artifacts without compromising pretrained representations. To address these limitations, a deepfake detection framework named FMSD is proposed, which integrates forgery-aware layer masking with multi-artifact subspace decomposition. Specifically, forgery-aware layer masking evaluates the bias-variance characteristics of layer-wise gradients to identify forgery-sensitive layers, thereby selectively updating them while reducing unnecessary disturbance to pretrained representations. Building upon this, multi-artifact subspace decomposition further decomposes the selected layer weights via Singular Value Decomposition (SVD) into a semantic subspace and multiple learnable artifact subspaces. These subspaces are optimized to capture heterogeneous and complementary forgery artifacts, enabling effective modeling of diverse forgery patterns while preserving pretrained semantic representations. Furthermore, orthogonality and spectral consistency constraints are imposed to regularize the artifact subspaces, reducing redundancy across them while preserving the overall spectral structure of pretrained weights. Extensive experiments demonstrate that our method achieves a frame AUC of up to 89.3\% and a video AUC of 93.0\% in cross-dataset evaluations, and outperforms State-of-the-Art (SOTA) methods by 4.5\% in whole synthetic face, 4.1\% in real-world settings, and 0.8\% in cross-manipulation.
\end{abstract}

\begin{IEEEkeywords}
Deepfake detection, Bias-variance ratio, Singular value decomposition, Cross-dataset generalization
\end{IEEEkeywords}

\section{Introduction}
\IEEEPARstart{I}{n} recent years, with the rapid advancement of generative artificial intelligence technologies~\cite{Karras2018ASG,J2021GenerativeAN,Rombach2021DiffusionModels,NEURIPS2020_4c5bcfec}, deepfake techniques have been able to manipulate facial regions with extremely high realism. The resulting forged images and videos can spread rapidly on social media platforms, posing serious threats to personal privacy and exerting profound impacts on the public opinion environment and social trust systems~\cite{Chesney2018DeepFA,Lee2023DeepfakesPS,busacca2023deepfake,kietzmann2020deepfakes,Vaccari2020DeepfakesAD,Wang2022DeepfakeDA}. In the face of continuously emerging forgery techniques, how to construct deepfake detection models that remain robust and highly generalizable under unknown forgery scenarios has become one of the core challenges in the field of digital media forensics.

\begin{figure}[htbp]
\centering
\includegraphics[width=0.5\textwidth]{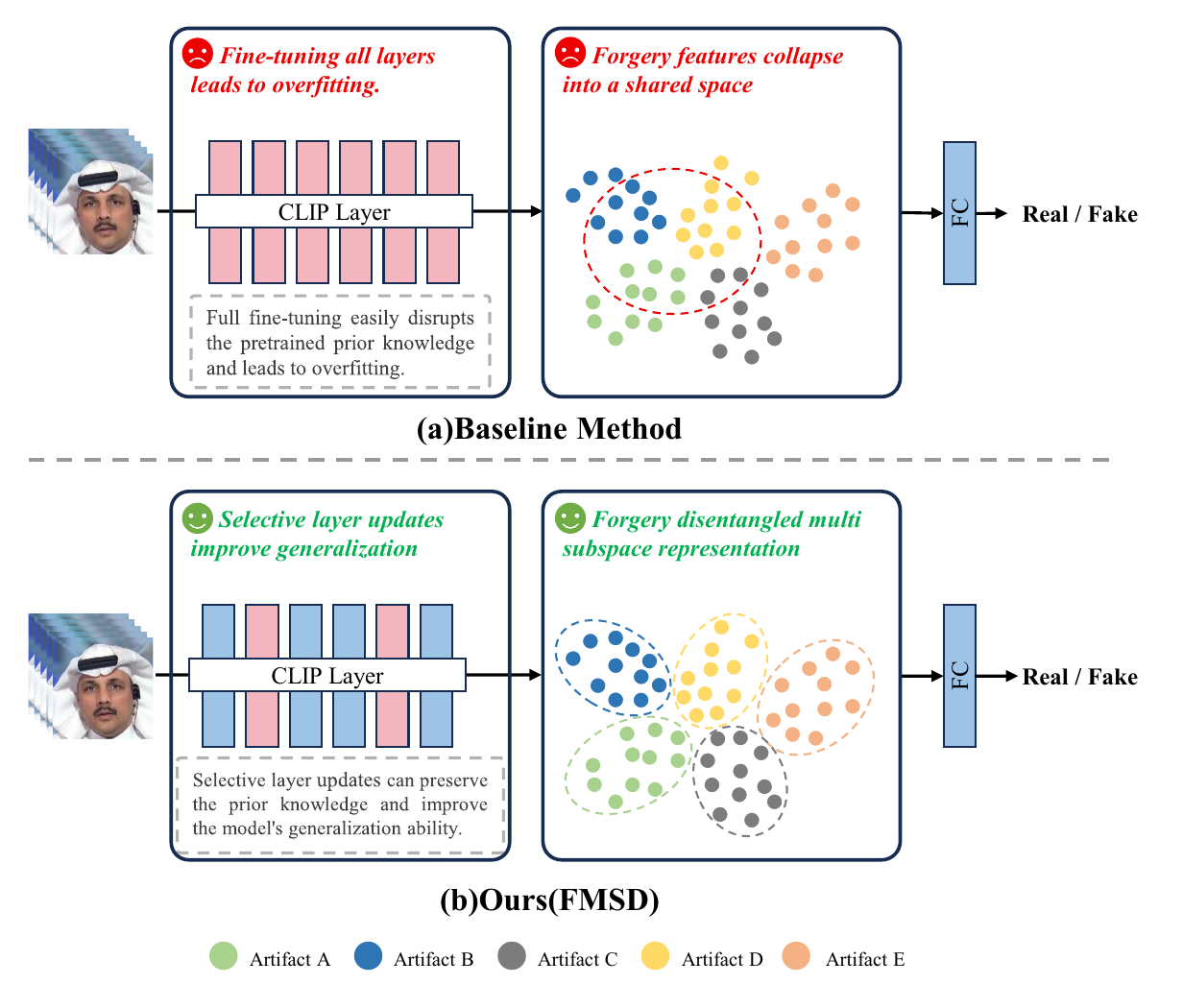}
\caption{Intuitive comparison between traditional deepfake detection methods and the proposed FMSD framework. (a) Baseline methods employ full-parameter fine-tuning, which often disrupts the original semantic representations and induces overfitting. Meanwhile, the model learns forgery features within a single shared space, leading to feature distribution collapse and diminished generalization to unseen forgery types. (b) In contrast, FMSD preserves pretrained semantic representations via selective layer updates and decouples forgery features into multiple artifact subspaces. This strategy effectively mitigates overfitting and enhances cross-dataset generalization capability.}
\label{motivation}
\end{figure}

Benefiting from the strong representation capability of Convolutional Neural Networks (CNNs)~\cite{Chollet2016XceptionDL,Tan2019EfficientNet}, deep learning–based forgery detection methods have achieved remarkable performance under in-distribution training and testing conditions. However, these methods often rely on local artifact patterns specific to particular forgery algorithms or datasets, and their detection performance degrades significantly when the test distribution shifts. To alleviate such overfitting issues, existing studies mainly enhance model generalization from two perspectives: data synthesis and representation learning.

At the data synthesis level, related methods construct or expand forged samples to cover a broader forgery space, such as boundary blending simulation or latent space augmentation, thereby guiding models to learn discriminative cues shared across different forgery methods~\cite{Li2018FWA,Li2019FaceXF,Shiohara2022SBI,Yan2023LSDA,Yan2024STA}. At the representation learning level, researchers improve robustness to distribution shifts through approaches including spatial–frequency fusion~\cite{Miao2023F2TransHF,Wang2023SFDG,Wang2026MAPMambaMP}, disentangled learning~\cite{Yan2023UCF,Lin2024PreservingFG}, attention mechanisms~\cite{Nguyen2024LAANetLA}, reconstruction learning~\cite{Tian2024RAM}, and transfer learning~\cite{Cui2025ForensicsAdapter,Fu2025UDD,Wang2026FromST,Kong2024MoE-FFD,Yan2024OrthogonalSD}.

However, as illustrated in Fig.~\ref{motivation}(a), 
existing deepfake detection methods typically fine-tune the entire network or most of its layers and model forged samples from diverse sources and manipulation methods within a shared feature space. This strategy suffers from two key limitations. First, during the transfer from general visual representations to the deepfake detection task, different network layers exhibit significant discrepancies in their learning dynamics. Updating all layers can easily disrupt the semantic representations learned during pretraining and induce overfitting, thereby compromising the model's generalization ability~\cite{Geirhos2020ShortcutLI,Yosinski2014HowTA,Li2016LearningWF,Li2023ExploreTE,Li2025PeerIY,Li2023APA}. Second, forgeries generated by different manipulation methods often exhibit highly heterogeneous artifact patterns. Forcing these diverse patterns into a shared feature space requires the model to simultaneously fit multiple fine-grained artifact features.Consequently, feature representations tend to collapse onto a few easier-to-fit forgery patterns, leading to feature distribution collapse. This severely weakens the model's ability to represent other forgery types, particularly unseen ones.

To address these challenges, a deepfake detection framework is proposed by integrating forgery-aware layer masking with multi-artifact subspace decomposition, with its core intuition illustrated in Fig.~\ref{motivation}(b). Specifically, FMSD introduces forgery-aware layer masking to adaptively regulate parameter updates across different network layers during fine-tuning. This module evaluates the bias-variance characteristics of layer-wise gradients to identify forgery-sensitive layers, thereby selectively updating them while reducing unnecessary disturbance to pretrained semantic representations. Building upon this, FMSD further performs multi-artifact subspace decomposition on the selected layers, decomposing their weights into a semantic subspace and multiple learnable artifact subspaces. These subspaces are designed to capture diverse forgery artifacts arising from different manipulation methods. Unlike conventional methods that rely on unified modeling within a single shared space, FMSD models heterogeneous forgery patterns from different sources in a more decoupled manner, thereby improving generalization to unseen forgery types. The main contributions of this paper are summarized as follows:

\begin{enumerate}
\item A generalizable deepfake detection framework is proposed, which integrated forgery-aware layer masking and multi-artifact subspace decomposition. Through this framework, only forgery-aware layers are selectively updated to enhance generalization, while diverse forgery artifacts are effectively disentangled and captured.

\item A forgery-aware layer masking strategy is presented, which identifies forgery-sensitive layers by evaluating the bias--variance characteristics of layer-wise gradients, thereby selectively updating layers that are more sensitive to forgery artifacts while preserving pretrained semantic representations.

\item A multi-artifact subspace decomposition mechanism is proposed that decomposes the selected layer weights into a semantic subspace and multiple learnable artifact subspaces, enabling decoupled modeling of heterogeneous forgery artifacts generated by different manipulation methods and mitigating feature distribution collapse in a single shared space.

\item Extensive experiments under five standard evaluation protocols validate the effectiveness of FMSD on multiple mainstream deepfake detection benchmarks. FMSD achieves 89.3\%/93.0\% frame/video AUC in cross-dataset evaluation. Moreover, in whole synthetic face, real-world settings, and cross-manipulation, it outperforms SOTA methods by 4.5\%, 4.1\%, and 0.8\% AUCs, respectively, highlighting its good generalization to unseen forgery scenarios.
\end{enumerate}

\section{Related Work}

\subsection{Deepfake Generation Methods}
Deepfake generation methods primarily manipulate facial identity, expression, or motion to produce forged content in real face images or videos. According to different generation methods, existing approaches can be broadly categorized into face replacement~\cite{2021FaceSwap,DeepFakes2020,Li2019CDF} and face reenactment~\cite{Thies2016Face2Face,thies2019NeuralTexture}. Face replacement typically achieves identity transfer and swapping through autoencoder-based or graphics-based techniques~\cite{2021FaceSwap,DeepFakes2020,Li2019CDF}. In contrast, face reenactment preserves the target identity while driving facial expressions or motions from a source video onto the target face~\cite{Thies2016Face2Face,thies2019NeuralTexture}. Although these deepfake methods differ in generation objectives, they generally involve joint modeling of appearance, geometric alignment, and temporal motion. Thus, the generated outputs often introduce multiple types of forgery artifacts~\cite{Li2019FaceXF,Rssler2019FF++,Chai2020WhatMF}. These artifacts exhibit significant variations in spatial distribution and representational levels, giving rise to the characteristic multi-artifact nature of face forgeries.

\subsection{Data Synthesis-Based Deepfake Detection}
In deepfake detection research, data synthesis–based training strategies are widely regarded as effective approaches. Such methods construct or mix synthetic samples to guide models toward learning discriminative cues shared by deepfakes, such as discontinuities along face blending boundaries and statistical inconsistencies between inner and outer facial regions, thereby improving generalization to unseen forgery methods. Early work, FWA~\cite{Li2018FWA}, adopts a self-blending strategy by applying image transformations such as downsampling to facial regions followed by inverse geometric mapping, in order to simulate warping artifacts introduced during deepfake generation and encourage the model to focus on abnormal patterns caused by geometric distortions. Subsequently, Face X-Ray~\cite{Li2019FaceXF} explicitly constructs blending boundaries to guide detectors to directly learn edge inconsistencies between real and forged regions. SBI~\cite{Shiohara2022SBI} performs face swapping using images of the same identity, significantly improving the realism of synthetic data while preserving identity consistency, thus reducing the distribution gap between simple synthetic samples and real forgeries. Building on this, SLADD~\cite{Chen2022SLADD} combines self-supervised learning with adversarial augmentation to dynamically generate more challenging forged samples, continuously exposing the model to diverse and evolving forgery patterns during training and enhancing its adaptability to complex forgery scenarios. More recently, StA~\cite{Yan2024STA} extends the perspective of data synthesis from the image level to the video level by simulating temporally varying facial feature drift through video-level data blending and introducing spatiotemporal adapters to explicitly model such dynamics. Beyond RGB appearance–based synthesis strategies, LSDA~~\cite{Yan2023LSDA} further explores the latent space by modeling variations of forgery features within latent representations and their relationships, effectively expanding the distribution of forged samples and providing richer training signals for deepfake detection.

\subsection{Representation Learning-Based Deepfake Detection}
Beyond data synthesis–based strategies, representation learning–based deepfake detection methods constitute another major research direction. These approaches do not rely on explicitly constructing forged samples; instead, they learn discriminative representations that capture intrinsic inconsistencies of deepfakes through feature space modeling, thereby improving generalization across forgery algorithms, compression settings, and demographic variations. Toward this goal, existing studies explore multiple perspectives, including spatial–frequency fusion, feature disentanglement, attention mechanisms, reconstruction consistency, and transfer learning. In spatial–frequency fusion, SFDG~\cite{Wang2023SFDG} jointly models spatial texture cues and frequency-domain statistical features, enhancing sensitivity to high-frequency artifacts and compression distortions. F2Trans~\cite{Miao2023F2TransHF} further integrates frequency-domain modeling with fine-grained Transformer architectures to capture subtle differences among high-frequency artifacts, thereby improving discriminative capability. UCF~\cite{Yan2023UCF} achieves feature disentanglement via unsupervised contrastive constraints, enabling the model to learn common representations shared across different forgery algorithms. 

From a fairness generalization perspective, FG~\cite{Lin2024PreservingFG} introduces feature disentanglement and loss flattening strategies to alleviate performance bias across demographic groups and domains. In artifact modeling, LAA-Net~\cite{Nguyen2024LAANetLA} employs localized artifact attention to guide the model toward key forged regions and combines it with a multi-task learning framework, endowing the detector with stronger adaptability across varying compression and quality conditions. In contrast to discriminative learning, RAM~\cite{Tian2024RAM} adopts a reconstruction-based paradigm that detects forgeries by reconstructing real appearances and measuring reconstruction consistency, thereby identifying forged samples from the perspective of generative consistency. In recent years, with the rise of large-scale pretrained models, transfer learning has become an important means of enhancing cross-domain generalization. FA~\cite{Cui2025ForensicsAdapter} introduces lightweight adapter modules on top of a frozen CLIP visual encoder to enhance discrimination of forged regions while preserving semantic consistency. UDD~\cite{Fu2025UDD} weakens positional and content biases by randomly shuffling and mixing Transformer tokens, encouraging the model to learn more robust, forgery-invariant representations. In addition, MoE-FFD~\cite{Kong2024MoE-FFD} and Effort~\cite{Yan2024OrthogonalSD} leverage feature decomposition and subspace modeling to mitigate overfitting while jointly preserving semantic information and learning forgery patterns.

\section{Methodology}

\begin{figure*}[htbp]
    \centering
    \includegraphics[width=0.9\textwidth]{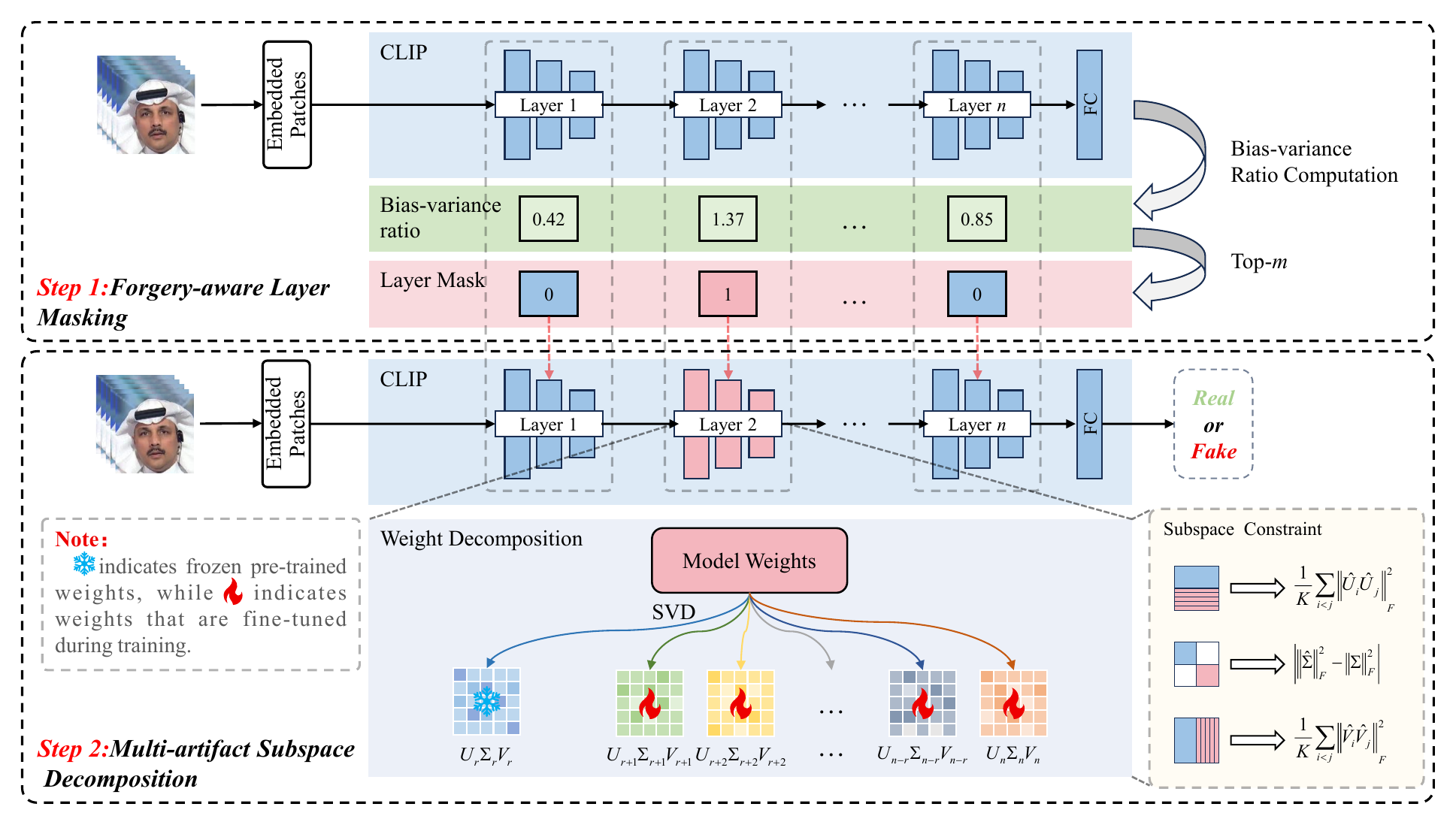}
    \caption{Overview of the FMSD framework. The model is initialized with pretrained weights and consists of two key steps: Step 1) \textit{Forgery-aware Layer Masking}, which generates layer-wise masks based on the bias--variance characteristics of layer-wise gradients to determine which layers should be updated; Step 2) \textit{Multi-Artifact Subspace Decomposition}, which decomposes the selected layer weights via SVD into a semantic subspace and multiple artifact subspaces, and optimizes them jointly with orthogonality and spectral consistency constraints.}
    \label{framework}
\end{figure*}

This paper proposes a new deepfake detection framework named FMSD. As illustrated in Fig.~\ref{framework}, FMSD comprises two key components: forgery-aware layer masking and multi-artifact subspace dcomposition. Forgery-aware layer masking evaluates the bias--variance characteristics of layer-wise gradients to identify forgery-sensitive layers for selective updating, thereby reducing unnecessary disturbance to pretrained semantic representations. Based on the selected layers, multi-artifact subspace decomposition further decomposes their weights into a semantic subspace and multiple learnable artifact subspaces to model diverse forgery artifacts in a more decoupled manner. In addition, orthogonality and spectral consistency constraints are introduced to regularize the artifact subspaces and enhance the discriminative capability of the model. The details of each component are presented in the following subsections.

\subsection{Forgery-aware Layer Masking}\label{3a}
In FMSD, a forgery-aware layer masking mechanism is proposed to generate layer-wise masks that determine whether each layer participates in parameter updates during fine-tuning. Specifically, let $B$ denote a mini-batch sampled from the training data. The parameter update of the $l$-th layer at iteration $t$ is formulated as:
\begin{equation}
\boldsymbol{\theta}^{(l,t)} \leftarrow \boldsymbol{\theta}^{(l,t-1)} + M^{(l,t)} \boldsymbol{d}^{(l,t)},
\end{equation}
where $\boldsymbol{\theta}^{(l,t)}$ and $\boldsymbol{\theta}^{(l,t-1)}$ denote the parameters of the $l$-th layer at iterations $t$ and $t-1$, respectively. $M^{(l,t)} \in \{0,1\}$ is a forgery-aware layer-wise mask indicating whether the $l$-th layer participates in parameter updates at iteration $t$, and $\boldsymbol{d}^{(l,t)}$ is the update direction given by standard gradient descent:

\begin{equation}
\boldsymbol{d}^{(l,t)} = -\eta \boldsymbol{g}^{(l,t)},
\end{equation}
where $\eta$ is the learning rate and $\boldsymbol{g}^{(l,t)} = \nabla_{\boldsymbol{\theta}^{(l)}} \mathcal{L}(\boldsymbol{\theta}^{(l,t)})$ is the gradient of the $l$-th layer at iteration $t$.

Next, in order to determine $M^{(l,t)}$, we draw inspiration from the bias-variance analysis framework in~\cite{Tomita2024ASF} and evaluate the bias-variance characteristics of each trainable layer, retaining only the top-$m$ layers for updating at each iteration. The layer-wise mask is then defined as:
\begin{equation}
M^{(l,t)} =
\begin{cases}
1, & l \in \mathrm{Top}\text{-}m(\mathrm{BVR}^{(l,t)}),\\
0, & \text{otherwise}.
\end{cases}
\end{equation}
where $\mathrm{BVR}^{(l,t)}$ denotes the bias-variance ratio of the $l$-th layer at iteration $t$. Intuitively, a larger $\mathrm{BVR}^{(l,t)}$ suggests that the gradients of the corresponding layer contain more reliable task-relevant information and are less dominated by stochastic fluctuations. In deepfake detection, such information is more likely to reflect forgery-related cues. Therefore, layers with larger $\mathrm{BVR}^{(l,t)}$ are regarded as more suitable for selective updating. The bias-variance ratio is defined as:

\begin{equation}
\mathrm{BVR}^{(l,t)} = \frac{b^{(l,t)}}{v^{(l,t)} + \epsilon},
\end{equation}
where $\epsilon$ is a small constant introduced to avoid numerical instability. $b^{(l,t)}$ and $v^{(l,t)}$ denote the bias and variance terms of the gradient in the $l$-th layer at iteration $t$, respectively, and are ideally defined as:
\begin{equation}
b^{(l,t)} = \left\| \mathbb{E}[\boldsymbol{g}^{(l,t)}] \right\|^2,
\qquad
v^{(l,t)} = \mathrm{tr}\!\left(\mathbb{V}[\boldsymbol{g}^{(l,t)}]\right).
\end{equation}

Notice that the bias term $b^{(l,t)}$ characterizes the consistent component of the gradient, whereas the variance term $v^{(l,t)}$ measures its stochastic fluctuation. Since these quantities cannot be computed directly under the true data distribution, this paper approximates them using the first- and second-order moments of the gradient. Accordingly, the practical form of $\mathrm{BVR}^{(l,t)}$ is written as:
\begin{equation}
\mathrm{BVR}^{(l,t)} =
\frac{\sum_i \left(\mu_i^{(l,t)}\right)^2}
{\sum_i \left(\sigma_i^{(l,t)} - \left(\mu_i^{(l,t)}\right)^2\right) + \epsilon},
\end{equation}
where $\mu_i^{(l,t)}$ and $\sigma_i^{(l,t)}$ denote the estimated first- and second-order moments of the $i$-th gradient element in the $l$-th layer at iteration $t$, respectively. These statistics are updated via exponential moving average as:
\begin{align}
\mu_i^{(l,t)} &= \alpha \mu_i^{(l,t-1)} + (1-\alpha) g_i^{(l,t)}, \\
\sigma_i^{(l,t)} &= \alpha \sigma_i^{(l,t-1)} + (1-\alpha)\left(g_i^{(l,t)}\right)^2,
\end{align}
where $\alpha \in [0,1)$ is the exponential smoothing coefficient.

\subsection{Multi-Artifact Subspace Decomposition}\label{3b}
The forgery-aware layer masking in Section~\ref{3a} determines which layers participate in parameter updates. Based on the selected layers, multi-artifact subspace decomposition is further proposed to determine how their weights are parameterized during adaptation. Specifically, for each selected layer, Singular Value Decomposition (SVD) is performed on its pretrained weight matrix and decompose the parameter space into a semantic subspace and multiple artifact subspaces.

Let $W \in \mathbb{R}^{d_{in} \times d_{out}}$ denote the weight matrix of a selected layer in the pretrained visual backbone. Applying singular value decomposition yields:
\begin{equation}
W = U \Sigma V^\top ,
\end{equation}
where $\Sigma = \mathrm{diag}(\sigma_1,\sigma_2,\ldots,\sigma_R)$ contains the singular values sorted in descending order, and $U$ and $V$ represent the corresponding left and right singular vector matrices, respectively.

The top-$r$ singular components are retrained to construct a rank-$r$ low-rank approximation of the weight matrix:
\begin{equation}
W_r = U_r \Sigma_r V_r^\top ,
\end{equation}
where $U_r \in \mathbb{R}^{d_{in} \times r}$, $\Sigma_r \in \mathbb{R}^{r \times r}$, and $V_r \in \mathbb{R}^{d_{out} \times r}$. This low-rank approximation captures the principal structure of the pretrained weight matrix and serves as the semantic subspace that preserves the primary knowledge of the pretrained model. During training, $W_r$ remains frozen to maintain pretrained semantic representations.

The residual component between the original pretrained weights and their low-rank approximation is defined as:
\begin{equation}
\Delta W = W - W_r .
\end{equation}

As the residual component $\Delta W$ complements the frozen semantic subspace, it provides a flexible parameter space for modeling task-specific variations, which is particularly suitable for learning forgery-related features.

However, representing all forgery artifacts within a single residual space is often insufficient to effectively model the diverse artifact patterns generated by different manipulation methods. When multiple types of artifact are forced to share the same representation space, feature competition and representation interference may occur, which limits the expressive capacity of the model. Therefore, multiple artifact subspaces are introduced to capture diverse forgery artifacts. Specifically, a representation subspace can be spanned by a set of basis vectors. Let $\mathcal{B}_k=\{b_{k,1},\ldots,b_{k,r_k}\}$ denote a set of basis vectors. The corresponding representation subspace is:
\begin{equation}
\mathcal{S}_k = \mathrm{span}(\mathcal{B}_k).
\end{equation}

When multiple sets of basis vectors are introduced into the residual space, the representational capacity of the model is expanded into a combination of multiple subspaces:
\begin{equation}
\mathcal{S} = \sum_{k=1}^{K} \mathcal{S}_k .
\end{equation}

By constructing multiple representation subspaces, the model can capture forgery artifact patterns with similar statistical properties in different subspaces. This enables the decoupled modeling of heterogeneous artifacts and improves the overall representational capacity of the model. By incorporating the above formulation into the residual space $\Delta W$, the residual space is further parameterized into $K$ independently optimizable artifact subspaces:
\begin{equation}
\Delta W = \sum_{k=1}^{K} U_k \Sigma_k V_k^\top ,
\end{equation}
where $U_k \in \mathbb{R}^{d_{in}\times r_k}$, $\Sigma_k \in \mathbb{R}^{r_k\times r_k}$, and $V_k \in \mathbb{R}^{d_{out}\times r_k}$. Each subspace is encouraged to capture complementary forgery artifact patterns. Through this multi-subspace decomposition, forgery features originating from diverse sources or manipulation methods are modeled in their respective subspaces, thereby mitigating mutual interference among heterogeneous artifact patterns.

During training, only the parameters within the residual space are updated, while the semantic subspace $W_r$ serves as a frozen anchor. Consequently, the effective weight used for forward propagation is given by:
\begin{equation}
\hat{W} = W_r + \Delta \hat{W}.
\end{equation}

However, subspace decomposition alone does not strictly guarantee independence among different artifact subspaces. To further enhance subspace separability and minimize representation overlap, additional subspace constraints are introduced during training, as formulated in Section~\ref{3c}.

\subsection{Objective Function}\label{3c}

\noindent\textbf{Subspace Orthogonality Constraint.}
To mitigate representation redundancy among different artifact subspaces, an orthogonality constraint that encourages distinct subspaces is introduced to learn complementary forgery features. Let the weight component of the $k$-th artifact subspace be represented via Singular Value Decomposition (SVD) as $(U_k,\Sigma_k,V_k)$, where $U_k \in \mathbb{R}^{d_{\text{in}}\times r_k}$, $\Sigma_k \in \mathbb{R}^{r_k\times r_k}$, and $V_k \in \mathbb{R}^{d_{\text{out}}\times r_k}$ represent the left singular vectors, singular values, and right singular vectors, respectively.

For any pair of distinct artifact subspaces $(i,j)$, we impose orthogonality constraints on both the left and right singular vector spaces, formulated as:
\begin{equation}
\mathcal{L}_{\text{orth}} =
\frac{2}{K(K-1)}
\sum_{i<j}
\Big(
\left\| V_i^\top V_j \right\|_F^2
+
\left\| U_i^\top U_j \right\|_F^2
\Big),
\end{equation}
where $K$ denotes the total number of artifact subspaces and $\|\cdot\|_F$ represents the Frobenius norm.

\noindent\textbf{Spectral Consistency Constraint.}
During multi-artifact subspace fine-tuning, the absence of additional constraints may allow the model to alter the singular value distribution, thereby shifting the overall spectral structure of the weight matrix and disrupting the stable semantic representations encoded in the pretrained model.

Given that the squared Frobenius norm satisfies $\|W\|_F^2=\sum_i \sigma_i^2$ (i.e., the sum of squared singular values), it serves as an effective measure of the weight matrix's overall spectral structure. 
Leveraging this property, a spectral consistency constraint is introduced to constrain the deviation in spectral structure between the pretrained and fine-tuned weights:
\begin{equation}
\mathcal{L}_{\text{spec}} =
\left| \|\hat{W}\|_F^2 - \|W\|_F^2 \right|,
\end{equation}
where $\hat{W}$ denotes the weight matrix at the current iteration and $W$ represents the corresponding pretrained weight matrix.

\noindent\textbf{Classification Loss.}
For deepfake detection, the standard cross-entropy loss is employed as the classification objective:
\begin{equation}
\mathcal{L}_{\text{cls}} =
-\frac{1}{N}\sum_{i=1}^{N}
\left(
y_i \log p_i +
(1-y_i)\log(1-p_i)
\right),
\end{equation}
where $y_i \in \{0,1\}$ denotes the ground-truth label of the $i$-th sample, $p_i$ is the predicted probability of the sample being forged, and $N$ is the mini-batch size.

\noindent\textbf{Overall Objective.}
The final optimization objective is:
\begin{equation}
\mathcal{L}_{\text{total}} =
\mathcal{L}_{\text{cls}}
+
\lambda_1 \frac{1}{n}\sum_{i=1}^{n}\mathcal{L}_{\text{orth}}^{(i)}
+
\lambda_2 \frac{1}{n}\sum_{i=1}^{n}\mathcal{L}_{\text{spec}}^{(i)},
\end{equation}
where $\lambda_1$ and $\lambda_2$ are weighting coefficients balancing the structural constraints and the classification objective, and $n$ denotes the number of network layers undergoing multi-artifact subspace fine-tuning.

\section{Experiments}
\subsection{Experimental Setup}
\noindent\textbf{Implementation Details.}
All experiments are conducted on a single NVIDIA RTX 4090 GPU using PyTorch 2.1.2. The pretrained CLIP ViT-L/14~\cite{Radford2021CLIP} is adoptd as the visual backbone. Following the default evaluation protocol of DeepfakeBench~\cite{Yan2023DeepfakeBenchAC}, FaceForensics++ (FF++)~\cite{Rssler2019FF++} is used for training. FF++ includes four manipulation subsets: DeepFakes (DF), Face2Face (F2F), FaceSwap (FS), and NeuralTextures (NT).
In FMSD, Forgery-aware Layer Masking (FLM) is applied to the self-attention linear layers of the backbone to generate layer-wise masks, and Multi-Artifact Subspace Decomposition (MASD) is further performed on the selected layers, while all other parameters remain frozen to preserve pretrained semantic representations. For each selected linear layer, we set the number of artifact subspaces to $K=5$ and the rank of each artifact subspace to $r_k=1$, yielding a semantic subspace rank of $r=1024-K\times r_k=1019$. The model is optimized using Adam with an initial learning rate of $2\times10^{-4}$, a batch size of 32, and a total of 10 training epochs. For both training and testing, 32 frames are uniformly sampled from each video. The training objective consists of the cross-entropy loss and two subspace regularization terms, namely the orthogonality constraint and the spectral consistency constraint, with trade-off weights $\lambda_1=1.0$ and $\lambda_2=1.0$, respectively.

\noindent\textbf{Evaluation Metrics.}
By default, the frame-level area under the ROC curve (AUC) is used as the primary metric for comparisons with prior work. For completeness and consistency with existing literature, video-level AUC is additionally reported. Accuracy (Acc.) and Equal Error Rate (EER) are further included as complementary evaluation metrics.

\noindent\textbf{Evaluation Protocols and Datasets.}
A comprehensive evaluation of our model is conducted under five standard evaluation protocols: \textbf{Protocol-1}: cross-dataset evaluation; \textbf{Protocol-2}: cross-manipulation evaluation within the FF++ data domain; \textbf{Protocol-3}: detection of entire synthetic faces; \textbf{Protocol-4}: detection in real-world scenarios; and \textbf{Protocol-5}: robustness evaluation under different distortion types and levels within the FF++ domain. Unless otherwise specified, all models are trained on FaceForensics++ (FF++) and evaluated according to the above protocols. FF++ provides multiple compression settings; following common practice in prior work, the c23 version is adopted for training.

In \textbf{Protocol-1}, generalization in the cross-dataset setting is assessed. The test datasets include Celeb-DF-v2 (CDF-v2)~\cite{Li2019CDF}, DFDC preview (DFDCP)~\cite{Dolhansky2019DFDCP}, Deepfake Detection Challenge (DFDC)~\cite{Dolhansky2020DFDC}, and DeepFakeDetection (DFD)~\cite{Google2019DFD}.
In \textbf{Protocol-2}, cross-manipulation generalization is evaluated under the same data domain using the latest deepfake dataset DF40~\cite{Yan2024DF40TN}. DF40 is built within the FF++ domain and contains forged samples generated by eight manipulation methods, including UniFace~\cite{Xu2022DesigningOU}, BlendFace~\cite{Shiohara2023BlendFaceRI}, MobSwap~\cite{Xu2022MobileFaceSwapAL}, e4s~\cite{Liu2022FineGrainedFS}, FaceDan~\cite{Rosberg2022FaceDancerPA}, FSGAN~\cite{Nirkin2019FSGANSA}, InSwap~\cite{Wang2023InSwapper}, and SimSwap~\cite{Chen2020SimSwapAE}. This setting introduces variations in manipulation techniques while keeping the data domain fixed, thereby enabling a more precise assessment of a detector's ability to generalize to unseen forgery methods.
In \textbf{Protocol-3}, DF40~\cite{Yan2024DF40TN} is also leveraged to evaluate the detection of entire synthetic faces. Specifically, we consider synthetic-face data generated by representative GAN and diffusion models, including VQGAN~\cite{Esser2020TamingTF}, StyleGAN2~\cite{Karras2019AnalyzingAI}, Stable Diffusion v2.1 (sd2.1)~\cite{Rombach2021DiffusionModels}, RDDM~\cite{Liu2023ResidualDD}, and DDIM~\cite{Song2020DenoisingDI}. Entire-face synthesis further stresses the model's generalization to fully synthesized scenarios.
In \textbf{Protocol-4}, our model is evaluated on two recent datasets, MFFI~\cite{Miao2025MFFIMF} and DDL~\cite{Miao2025DDLAL}. These datasets contain a broader diversity of forgery techniques and facial scenarios, serving as more challenging benchmarks to assess the model’s detection capability in real-world settings.
In \textbf{Protocol-5}, distortions of different types and severities on FF++ are introduced to evaluate robustness, verifying the stability of our method under challenging real-world conditions.

It is worth noting that the first four protocols are designed with different evaluation emphases. \textbf{Protocol-1} and \textbf{Protocol-2} follow standard benchmark settings, where a relatively large number of prior methods and publicly reported AUC results are available. Accordingly, we mainly adopt the broad method comparison using AUC as the primary metric. In contrast, \textbf{Protocol-3} and \textbf{Protocol-4} target more recent and less-explored scenarios, for which publicly available results remain limited. Therefore, some representative baselines are reproduced under our unified implementation and evaluation pipeline, and multiple metrics are reported to provide a more comprehensive evaluation in challenging and realistic settings.

\subsection{Comparison with existing methods}
\noindent\textbf{Protocol 1: Cross-Dataset Evaluation.}
FMSD is compared with representative baselines covering multiple mainstream directions in deepfake detection, and adopt AUC as the primary metric because it is the most widely reported indicator in standard cross-dataset evaluation. To comprehensively assess cross-domain generalization, our method is further evaluated under both frame-level and video-level settings. The results are summarized as follows.

Table~\ref{frame_auc_comparison} reports the frame-level AUC comparison against 13 representative detectors. Overall, the proposed FMSD consistently achieves strong cross-domain generalization across all test datasets. In particular, FMSD attains frame-level AUCs of 84.4\% and 94.8\% on DFDC and DFD, respectively, outperforming all competing methods. On CDF-v2 and DFDCP, FMSD achieves 89.5\% and 88.4\%, respectively, ranking second and trailing FA~\cite{Cui2025ForensicsAdapter} by a small margin. Notably, FA introduces an additional mask-prediction branch to explicitly guide the model to focus on facial boundary regions, which can be advantageous for face-swapping manipulations. However, this design relies on extra pixel-level supervision, imposing stronger annotation requirements and thus limiting practical applicability. In contrast, FMSD does not require explicit mask supervision and can be trained solely with raw images, while still delivering competitive or superior performance across diverse forgery datasets, demonstrating robust generalization.

Table~\ref{video_auc_comparison} presents the video-level AUC comparison with 17 representative methods. FMSD achieves the best overall performance across all test datasets, with an average video-level AUC of 93.0\%. Compared with the strongest competitors, StA~\cite{Yan2024STA} and Effort~\cite{Yan2024OrthogonalSD}, FMSD improves the average performance by approximately 1.4\%, indicating stronger cross-domain generalization in realistic and challenging video forgery scenarios. It is worth noting that StA is a typical video-level detector that explicitly models inter-frame inconsistencies to capture temporal forgery cues, and thus can benefit from multi-frame inputs. In contrast, FMSD achieves superior overall performance without introducing additional temporal modeling modules, relying only on single-frame images to effectively capture forgery traces, which further validates the robustness and generality of our learned representations.

\begin{table*}[htbp]
\centering
\renewcommand{\arraystretch}{1.1}
\setlength{\tabcolsep}{5pt}
\caption{Cross-dataset evaluation results (\textbf{Frame-level AUC}). Methods marked with * are reproduced using publicly available models and weights, while the remaining results are quoted from~\cite{Yan2023DeepfakeBenchAC,Yan2023LSDA,wang2025idcnet,Kashiani2025FreqDebias}. \textbf{Bold} denotes the best performance, and \underline{underlined} denotes the second best.}
\begin{small}
\begin{tabular}{
l|
c|
>{\centering\arraybackslash}m{1.4cm}
>{\centering\arraybackslash}m{1.4cm}
>{\centering\arraybackslash}m{1.4cm}
>{\centering\arraybackslash}m{1.4cm}|
>{\centering\arraybackslash}m{1.4cm}}
\toprule
\textbf{Method} & \textbf{Venue} &
\textbf{CDF-v2} & \textbf{DFDCP} & \textbf{DFDC} & \textbf{DFD} & \textbf{Avg.} \\
\midrule
SRM~\cite{Luo2021SRM}        & CVPR 2021  & 0.755 & 0.741 & 0.700 & 0.812 & 0.752 \\
SPSL~\cite{Liu2021SPSL}      & CVPR 2021  & 0.765 & 0.741 & 0.704 & 0.812 & 0.756 \\
Recce~\cite{Cao2022Recce}    & CVPR 2022  & 0.732 & 0.734 & 0.713 & 0.812 & 0.748 \\
CORE*~\cite{Ni2022CORE}      & CVPRW 2022 & 0.743 & 0.734 & 0.705 & 0.802 & 0.746 \\
SLADD~\cite{Chen2022SLADD}   & CVPR 2022  & 0.740 & 0.753 & 0.717 & 0.809 & 0.755 \\
UCF~\cite{Yan2023UCF}        & ICCV 2023  & 0.753 & 0.759 & 0.719 & 0.807 & 0.760 \\
IID*~\cite{huang2023IID}     & CVPR 2023  & 0.769 & 0.762 & 0.695 & 0.793 & 0.755 \\
IDCNet~\cite{wang2025idcnet} & TIFS 2024  & 0.809 & 0.741 & 0.724 & 0.847 & 0.780 \\
ProDet*~\cite{cheng2024ProDet} & NeurIPS 2024 & 0.842 & 0.774 & 0.697 & 0.848 & 0.790 \\
LSDA~\cite{Yan2023LSDA}      & CVPR 2024  & 0.830 & 0.815 & 0.736 & 0.880 & 0.815 \\
Effort*~\cite{Yan2024OrthogonalSD} & ICML 2025 & 0.882 & 0.876 & 0.815 & 0.933 & 0.877 \\
FreqDebias~\cite{Kashiani2025FreqDebias} & CVPR 2025 & 0.836 & 0.824 & 0.741 & 0.868 & 0.817 \\
FA~\cite{Cui2025ForensicsAdapter} & CVPR 2025 & \textbf{0.899} & \textbf{0.890} & \underline{0.843} & \underline{0.933} & \underline{0.891} \\
\midrule
\textbf{FMSD} & Ours
& \underline{0.895} & \underline{0.884} & \textbf{0.844} & \textbf{0.948} & \textbf{0.893} \\
\bottomrule
\end{tabular}
\end{small}
\label{frame_auc_comparison}
\end{table*}

\begin{table*}[htbp]
\centering
\renewcommand{\arraystretch}{1.1}
\setlength{\tabcolsep}{5pt}
\caption{Cross-dataset evaluation results (\textbf{Video-level AUC}). Methods marked with * are reproduced using publicly available models and weights, while the remaining results are quoted from~\cite{Yan2023DeepfakeBenchAC,Fu2025UDD,Yan2024STA}. \textbf{Bold} denotes the best performance, and \underline{underlined} denotes the second best.}
\begin{small}
\begin{tabular}{
l|
c|
>{\centering\arraybackslash}m{1.4cm}
>{\centering\arraybackslash}m{1.4cm}
>{\centering\arraybackslash}m{1.4cm}
>{\centering\arraybackslash}m{1.4cm}|
>{\centering\arraybackslash}m{1.4cm}
}
\toprule
\textbf{Method} & \textbf{Venue} &
\textbf{CDF-v2} & \textbf{DFDCP} & \textbf{DFDC} & \textbf{DFD} & \textbf{Avg.} \\
\midrule
SRM~\cite{Luo2021SRM}   & CVPR 2021 & 0.840 & 0.728 & 0.695 & 0.885 & 0.787 \\
SPSL~\cite{Liu2021SPSL}  & CVPR 2021 & 0.799 & 0.770 & 0.724 & 0.871 & 0.791 \\
Recce~\cite{Cao2022Recce} & CVPR 2022 & 0.823 & 0.734 & 0.696 & 0.891 & 0.786 \\
CORE~\cite{Ni2022CORE}  & CVPRW 2022 & 0.809 & 0.720 & 0.721 & 0.882 & 0.783 \\
DCL~\cite{Sun2021DCL}   & AAAI 2022 & 0.882 & 0.769 & 0.750 & 0.921 & 0.831 \\
SLADD*~\cite{Chen2022SLADD} & CVPR 2022 & 0.837 & 0.756 & 0.772 & 0.904 & 0.817 \\
SBI~\cite{Shiohara2022SBI}   & CVPR 2022 & 0.886 & 0.848 & 0.717 & 0.827 & 0.820 \\
CFM~\cite{Luo2023CFM}   & TIFS 2023 & 0.897 & 0.802 & 0.706 & 0.952 & 0.839 \\
UCF~\cite{Yan2023UCF}   & ICCV 2023 & 0.837 & 0.770 & 0.742 & 0.867 & 0.804 \\
IID*~\cite{huang2023IID}   & CVPR 2023 & 0.787 & 0.717 & 0.709 & 0.828 & 0.760 \\
AltFreezing*~\cite{wang2023altfreezing} & CVPR 2023 & 0.801 & 0.667 & 0.640 & 0.718 & 0.707 \\
ProDet*~\cite{cheng2024ProDet} & NeurIPS 2024 & 0.926 & 0.828 & 0.725 & 0.901 & 0.845 \\
CDFA*~\cite{Lin2024CDFA}  & ECCV 2024 & 0.938 & 0.830 & 0.830 & 0.954 & 0.888 \\
LSDA~\cite{Yan2023LSDA}  & CVPR 2024 & 0.898 & 0.812 & 0.735 & 0.956 & 0.850 \\
UDD~\cite{Fu2025UDD}   & AAAI 2025 & 0.931 & 0.881 & 0.812 & 0.955 & 0.895 \\
StA~\cite{Yan2024STA}   & CVPR 2025 & \underline{0.947} & 0.909 & \underline{0.843} & 0.965 & \underline{0.916} \\
Effort*~\cite{Yan2024OrthogonalSD} & ICML 2025 & 0.944 & \underline{0.912} & 0.838 & \underline{0.973} & \underline{0.916} \\
\midrule
\textbf{FMSD} & Ours & \textbf{0.952} & \textbf{0.914} & \textbf{0.871} & \textbf{0.982} & \textbf{0.930} \\
\bottomrule
\end{tabular}
\end{small}
\label{video_auc_comparison}
\end{table*}

\noindent\textbf{Protocol 2: Cross-Manipulation Evaluation.}
Representative and competitive baselines are selected for generalizable deepfake detection, and adopt video-level AUC as the evaluation metric because it better reflects robustness to unseen forgery methods in practical scenarios. In practical deployment, detectors must often handle previously unseen manipulation types. To evaluate the generalization of FMSD across manipulation methods, cross-manipulation evaluation is conducted on DF40~\cite{Yan2024DF40TN} and compare it against 15 representative detectors. As shown in Table~\ref{Cross-manipulation evaluations}, FMSD achieves an AUC of 95.3\%, outperforming prior methods and indicating stronger robustness to variations in forgery techniques.

\begin{table*}[htbp]
\centering
\renewcommand{\arraystretch}{1.1}
\setlength{\tabcolsep}{1pt}
\caption{Cross-manipulation evaluation results (\textbf{Video-level AUC}). Methods marked with * are reproduced using publicly available models and weights, while the remaining results are quoted from~\cite{Yan2023DeepfakeBenchAC,Yan2024STA,Yan2024OrthogonalSD}. \textbf{Bold} denotes the best performance, and \underline{underlined} denotes the second best.}
\begin{small}
\begin{tabular}{
l|
c|
>{\centering\arraybackslash}m{1.5cm}
>{\centering\arraybackslash}m{1.5cm}
>{\centering\arraybackslash}m{1.5cm}
>{\centering\arraybackslash}m{1.5cm}
>{\centering\arraybackslash}m{1.5cm}
>{\centering\arraybackslash}m{1.5cm}
>{\centering\arraybackslash}m{1.5cm}
>{\centering\arraybackslash}m{1.5cm}|
>{\centering\arraybackslash}m{1cm}
}
\toprule
\textbf{Method} & \textbf{Venue} &
\textbf{UniFace} & \textbf{BlendFace} & \textbf{MobSwap} & \textbf{e4s} & \textbf{FaceDan} & \textbf{FSGAN} & \textbf{InSwap} & \textbf{SimSwap} & \textbf{Avg.} \\
\midrule
SRM~\cite{Luo2021SRM}  & CVPR 2021 & 0.749 & 0.704 & 0.779 & 0.704 & 0.659 & 0.772 & 0.793 & 0.694 & 0.732 \\
SPSL~\cite{Liu2021SPSL} & CVPR 2021 & 0.747 & 0.748 & 0.885 & 0.514 & 0.666 & 0.812 & 0.643 & 0.665 & 0.710 \\
Recce~\cite{Cao2022Recce} & CVPR 2022 & 0.898 & 0.832 & 0.925 & 0.683 & 0.848 & 0.949 & 0.848 & 0.768 & 0.844 \\
CORE~\cite{Ni2022CORE} & CVPRW 2022 & 0.871 & 0.843 & 0.959 & 0.679 & 0.774 & 0.958 & 0.855 & 0.724 & 0.833 \\
SLADD~\cite{Chen2022SLADD} & CVPR 2022 & 0.878 & 0.882 & 0.954 & 0.765 & 0.825 & 0.943 & 0.879 & 0.794 & 0.865 \\
SBI~\cite{Shiohara2022SBI} & CVPR 2022 & 0.724 & 0.891 & 0.952 & 0.750 & 0.594 & 0.803 & 0.712 & 0.701 & 0.766 \\
UCF~\cite{Yan2023UCF} & ICCV 2023 & 0.831 & 0.827 & 0.950 & 0.731 & 0.862 & 0.937 & 0.809 & 0.647 & 0.824 \\
IID~\cite{huang2023IID} & CVPR 2023 & 0.839 & 0.789 & 0.888 & 0.766 & 0.844 & 0.927 & 0.789 & 0.644 & 0.811 \\
AltFreezing~\cite{wang2023altfreezing} & CVPR 2023 & 0.947 & \textbf{0.951} & 0.851 & 0.605 & 0.836 & 0.958 & 0.927 & 0.797 & 0.859 \\
ProDet~\cite{cheng2024ProDet} & NeurIPS 2024 & 0.908 & \underline{0.929 } & 0.975 & 0.771 & 0.747 & 0.928 & 0.837 & 0.844 & 0.867 \\
CDFA~\cite{Lin2024CDFA} & ECCV 2024 & 0.762 & 0.756 & 0.823 & 0.631 & 0.803 & 0.942 & 0.772 & 0.757 & 0.781 \\
LSDA~\cite{Yan2023LSDA} & CVPR 2024 & 0.872 & 0.875 & 0.930 & 0.694 & 0.721 & 0.939 & 0.855 & 0.793 & 0.835 \\
Effort~\cite{Yan2024OrthogonalSD} & ICML 2025 & \textbf{0.962} & 0.873 & 0.953 & \textbf{0.983} & 0.926 & 0.957 & 0.936 & 0.926 & 0.940 \\
FA*~\cite{Cui2025ForensicsAdapter} & CVPR 2025 & 0.951 & 0.880 & \textbf{0.979} & 0.969 & \underline{0.946} & \textbf{0.975} & \textbf{0.949} & 0.913 & \underline{0.945} \\
StA~\cite{Yan2024STA} & CVPR 2025 & 0.960 & 0.906 & 0.946 & \underline{0.980} & 0.912 & 0.964 & 0.937 & \textbf{0.931} & 0.942 \\
\midrule
\textbf{FMSD} & Ours & \underline{0.961} & 0.915 & \textbf{0.979} & 0.976 & \textbf{0.962} & \underline{0.966} & \underline{0.942} & \underline{0.927} & \textbf{0.953} \\
\bottomrule
\end{tabular}
\end{small}
\label{Cross-manipulation evaluations}
\end{table*}

\noindent\textbf{Protocol 3: Entire Synthetic Face Evaluation.}
With the rapid advancement of generative models, both GANs and diffusion models can produce highly photorealistic fully synthesized faces, making performance in fully synthetic scenarios an important indicator of model generalization. Since this setting remains relatively underexplored in prior work, we report AUC, AP, and EER and reproduce several representative detectors under a unified evaluation pipeline for fair comparison. Five entire synthetic face datasets are considered, including VQGAN~\cite{Esser2020TamingTF}, StyleGAN2~\cite{Karras2019AnalyzingAI}, sd2.1~\cite{Rombach2021DiffusionModels}, RDDM~\cite{Liu2023ResidualDD}, and DDIM~\cite{Song2020DenoisingDI}. The results are reported in Table~\ref{detection of entire synthetic faces}. FMSD achieves the best or second-best performance across all datasets, with an average AUC of 94.6\%, an AP of 93.3\%, and an EER as low as 10.7\%, demonstrating strong generalization even in fully synthetic settings.

\begin{table*}[htbp]
\centering
\renewcommand{\arraystretch}{1.1}
\caption{Evaluation results on entire synthetic faces . Metrics are reported as AUC \textbar\ AP \textbar\ EER (frame-level). \textbf{Bold} denotes the best performance, and \underline{underlined} denotes the second best.}
\resizebox{\linewidth}{!}{\Large
\begin{tabular}{c|c|ccc|ccc|ccc|ccc|ccc|ccc}
\toprule
\multirow{2}{*}{\raisebox{-0.3em}{\textbf{Method}}} &
\multirow{2}{*}{\raisebox{-0.3em}{\textbf{Venue}}} &
\multicolumn{3}{c|}{\textbf{VQGAN}} &
\multicolumn{3}{c|}{\textbf{StyleGAN2}} &
\multicolumn{3}{c|}{\textbf{sd2.1}} &
\multicolumn{3}{c|}{\textbf{RDDM}} &
\multicolumn{3}{c|}{\textbf{DDIM}} &
\multicolumn{3}{c}{\textbf{Avg.}} \\
\cmidrule(lr){3-20}
& &
AUC & AP & EER &
AUC & AP & EER &
AUC & AP & EER &
AUC & AP & EER &
AUC & AP & EER &
AUC & AP & EER \\
\midrule
SRM~\cite{Luo2021SRM} & CVPR 2021
& 0.672 & 0.588 & 38.5
& 0.897 & 0.821 & 17.8
& 0.878 & 0.827 & 19.5
& 0.814 & 0.746 & 25.4
& 0.814 & 0.746 & 25.3
& 0.815 & 0.746 & 25.3 \\

SPSL~\cite{Liu2021SPSL} & CVPR 2021
& 0.650 & 0.574 & 39.6
& 0.854 & 0.829 & 23.3
& 0.702 & 0.577 & 35.1
& 0.382 & 0.363 & 60.9
& 0.806 & 0.769 & 27.0
& 0.679 & 0.622 & 37.2 \\

Recce~\cite{Cao2022Recce} & CVPR 2022
& 0.805 & 0.758 & 27.2
& 0.968 & 0.956 & 9.9
& 0.947 & 0.925 & 11.9
& 0.740 & 0.559 & 29.8
& 0.782 & 0.719 & 29.0
& 0.849 & 0.783 & 21.5 \\

UCF~\cite{Yan2023UCF} & ICCV 2023
& 0.772 & 0.719 & 30.3
& 0.914 & 0.906 & 16.7
& 0.939 & 0.922 & 12.6
& 0.712 & 0.535 & 33.2
& 0.838 & 0.808 & 24.6
& 0.835 & 0.778 & 23.5 \\

ProDet~\cite{cheng2024ProDet} & NIPS 2024
& 0.781 & 0.722 & 29.1
& 0.929 & 0.885 & 14.1
& \underline{0.971} & \underline{0.953} & \underline{7.9}
& 0.581 & 0.452 & 43.0
& \textbf{0.895} & \textbf{0.840} & \textbf{17.8}
& 0.831 & 0.771 & 22.4 \\

FA~\cite{Cui2025ForensicsAdapter} & CVPR 2025
& \underline{0.940} & \underline{0.939} & \underline{12.6}
& \textbf{0.997} & \textbf{0.996} & \textbf{2.1}
& 0.955 & 0.947 & 11.2
& \underline{0.903} & \underline{0.863} & \underline{17.0}
& 0.712 & 0.681 & 34.0
& \underline{0.901} & \underline{0.885} &\underline{ 15.4} \\

\midrule
FMSD & Ours
& \textbf{0.989} & \textbf{0.987} & \textbf{5.2}
& \underline{0.994} & \underline{0.993} & \underline{3.0}
& \textbf{0.989} & \textbf{0.987} & \textbf{5.2}
& \textbf{0.939} & \textbf{0.915} & \textbf{13.6}
& \underline{0.816} & \underline{0.785} & \underline{26.6}
& \textbf{0.946} & \textbf{0.933} & \textbf{10.7} \\
\bottomrule
\end{tabular}
}
\label{detection of entire synthetic faces}
\end{table*}

\noindent\textbf{Protocol 4: Detection in real-world scenarios.}
Since real-world deepfake detection is more challenging and remains relatively underexplored in prior work, we report multiple metrics and reproduce several representative detectors under a unified evaluation pipeline for fair comparison. Table~\ref{detection in real-world scenarios.} presents the detection results on the real-world datasets MFFI~\cite{Miao2025MFFIMF} and DDL~\cite{Miao2025DDLAL}. It can be observed that our method, FMSD, achieves the best or second-best performance on both datasets and consistently outperforms existing methods in terms of the overall average metrics, demonstrating stronger robustness and generalization capability in complex real-world scenarios.

\begin{table*}[t]
\centering
\caption{Evaluation results for detection in real-world scenarios (\textbf{Frame-level AUC}). The evaluation metrics are reported in the order of AUC \textbar\ ACC \textbar\ AP \textbar\ EER (frame level). \textbf{Bold} indicates the best results, while \underline{underlined} denotes the second-best results.}
\begin{small}
\begin{tabular}{c|c|cccc|cccc|cccc}
\toprule
\multirow[c]{2}{*}{\textbf{Method}} &
\multirow[c]{2}{*}{\textbf{Venue}} &
\multicolumn{4}{c|}{\textbf{MFFI}} &
\multicolumn{4}{c|}{\textbf{DDL}} &
\multicolumn{4}{c}{\textbf{Avg.}} \\
\cmidrule(lr){3-14}
& &
AUC & ACC & AP & EER &
AUC & ACC & AP & EER &
AUC & ACC & AP & EER \\
\midrule
SRM~\cite{Luo2021SRM} & CVPR 2021
& 0.679 & 0.620 & 0.730 & 37.5
& 0.503 & 0.560 & 0.806 & 53.0
& 0.591 & 0.590 & 0.768 & 45.2 \\

SPSL~\cite{Liu2021SPSL} & CVPR 2021
& 0.567 & 0.564 & 0.645 & 45.7
& 0.613 & 0.607 & 0.850 & 41.5
& 0.590 & 0.585 & 0.748 & 43.6 \\

Recce~\cite{Cao2022Recce} & CVPR 2022
& 0.641 & 0.559 & 0.691 & 39.1
& 0.530 & 0.431 & 0.821 & 50.8
& 0.585 & 0.495 & 0.756 & 45.0 \\

UCF~\cite{Yan2023UCF} & ICCV 2023
& 0.631 & 0.555 & 0.694 & 39.8
& 0.535 & 0.484 & 0.826 & 51.0
& 0.583 & 0.519 & 0.760 & 45.4 \\

ProDet~\cite{cheng2024ProDet} & NIPS 2024
& 0.616 & 0.594 & 0.674 & 42.0
& 0.643 & 0.678 & 0.849 & 38.6
& 0.630 & 0.636 & 0.761 & 40.3 \\

FA~\cite{Cui2025ForensicsAdapter} & CVPR 2025
& \textbf{0.765} & \underline{0.674} & \textbf{0.840} & \textbf{30.3}
& \underline{0.834} & \underline{0.715} & \underline{0.948} & \underline{24.7}
& \underline{0.800} & \underline{0.694} & \underline{0.894} & \underline{27.5} \\

\midrule
FMSD & Ours
& \underline{0.759} & \textbf{0.691} & \underline{0.832} & \underline{30.6}
& \textbf{0.922} & \textbf{0.838} & \textbf{0.977} & \textbf{14.3}
& \textbf{0.841} & \textbf{0.765} & \textbf{0.904} & \textbf{22.5} \\
\bottomrule
\end{tabular}
\end{small}
\label{detection in real-world scenarios.}
\end{table*}

\begin{figure*}[t]
    \centering
    \includegraphics[width=0.9\linewidth]{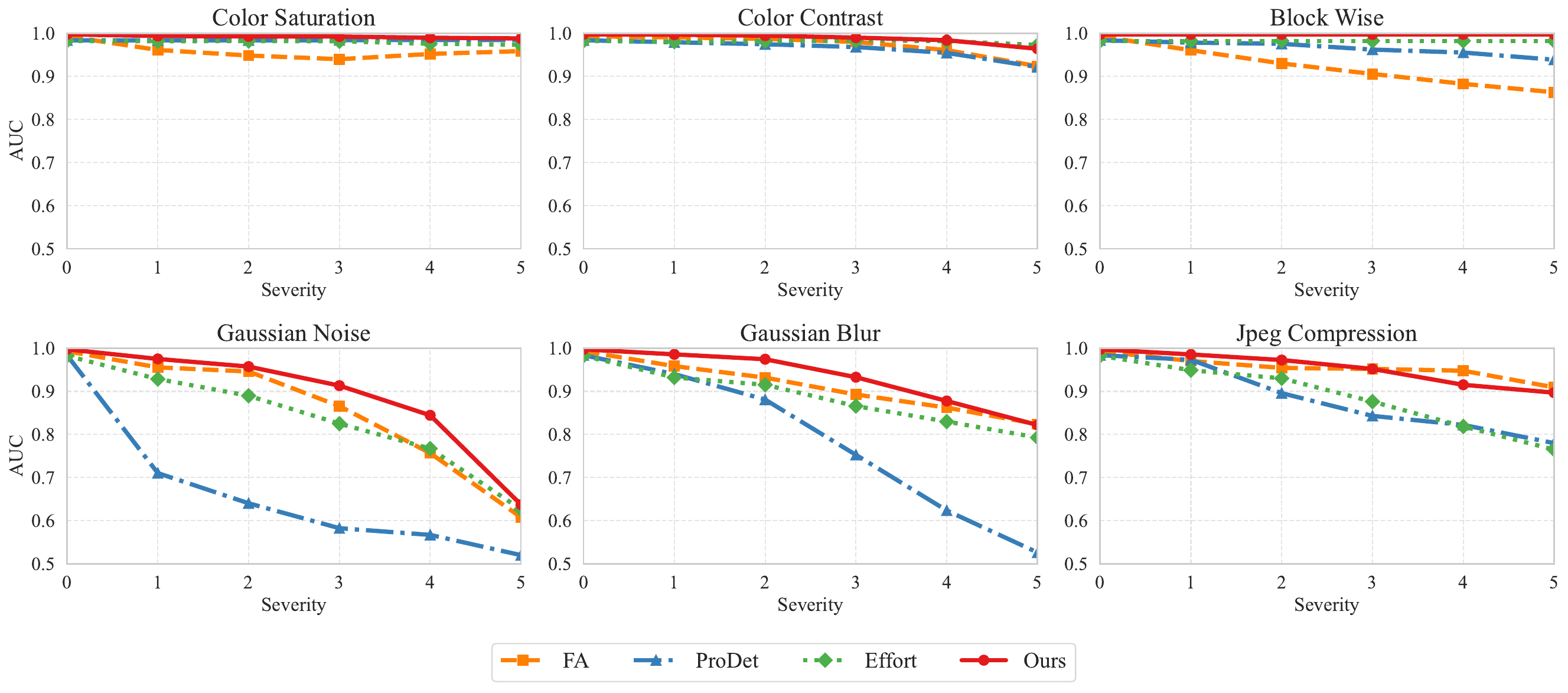}
    \caption{Robustness evaluation results (\textbf{Video-level AUC}). Comparison between our method and FA~\cite{Cui2025ForensicsAdapter}, ProDet~\cite{cheng2024ProDet}, and Effort~\cite{Yan2024OrthogonalSD} under six distortion types and five severity levels.}
    \label{robust_exp}
\end{figure*}

\noindent\textbf{Protocol 5: Robustness Evaluation.}
To assess robustness in real-world conditions, we adopt the perturbation protocol introduced in~\cite{Jiang2020DeeperForensics10AL}, which is widely used for robustness evaluation. Specifically, six representative distortions are introduced to the test videos, including color saturation shift, color contrast shift, block-wise artifacts, Gaussian blur, JPEG compression, and additive Gaussian noise. For each distortion, five severity levels are considered to systematically emulate quality degradations commonly encountered in practical scenarios.
Under this evaluation protocol, FMSD is compared with several recent and competitive methods, including FA~\cite{Cui2025ForensicsAdapter}, ProDet~\cite{cheng2024ProDet}, and Effort~\cite{Yan2024OrthogonalSD}. Video-level AUC is adopted to evaluate the stability of detection performance under different distortion types and severity levels. Fig.~\ref{robust_exp} reports the corresponding results.
As can be observed, our method maintains consistently stable detection performance across distortion types and severity levels, exhibiting stronger overall robustness than the competing methods, whose performance drops more noticeably under certain high-severity corruptions. These results further validate the applicability and reliability of our method in complex real-world environments.

\subsection{Ablation Studies and Analysis}
\noindent\textbf{Effects of MASD and FLM.}
To quantify the contribution of each core component, ablation studies are conducted on MASD and FLM. Specifically, one component is removed at a time while keeping the remaining architecture and training configuration unchanged to ensure fair comparisons. Area Under the Curve (AUC), Average Precision (AP), and Equal Error Rate (EER) are reported, with results summarized in Table~\ref{Method ablation}.
Note that our method modifies only the linear projections in the self-attention layers. Therefore, in the ablation setting without MASD, the selected self-attention linear layers is directly optimized without subspace decomposition, while freezing all other parameters to maintain comparability across settings. As shown in Table~\ref{Method ablation}, MASD yields the most substantial performance gain, suggesting that decomposing the selected layer weights into semantic and artifact subspaces effectively strengthens forgery representation learning. When MASD and FLM are jointly enabled, the model achieves the best performance across almost all metrics, validating the complementarity of the two components within a unified framework.
Moreover, in the two settings without MASD, the training schedule is fixed to 10 epochs for consistency. Under this configuration, the models have not fully converged and their performance continues to improve toward the end of training. This partially explains why enabling FLM alone can be slightly inferior to the variant without either component: the gap is mainly attributed to insufficient convergence under constrained trainable parameters rather than a limitation of the masking mechanism itself.

\begin{table*}[htbp]
\centering
\caption{Ablation study on the effectiveness of the proposed MASD and FLM strategies. All models are trained on FF++ (c23)~\cite{Rssler2019FF++} and evaluated in the cross-dataset setting. Metrics are reported as AUC \textbar\ AP \textbar\ EER (frame-level). The average performance across all datasets (Avg.) is additionally reported, and the best results are highlighted in \textbf{bold}.}
\begin{tabular}{c|c|ccc|ccc|ccc|ccc|ccc}
\toprule
\multirow{2}{*}{\raisebox{-0.3em}{\textbf{MASD}}} &
\multirow{2}{*}{\raisebox{-0.3em}{\textbf{FLM}}} &
\multicolumn{3}{c|}{\textbf{CDF-v2}} &
\multicolumn{3}{c|}{\textbf{DFDCP}} &
\multicolumn{3}{c|}{\textbf{DFDC}} &
\multicolumn{3}{c|}{\textbf{DFD}} &
\multicolumn{3}{c}{\textbf{Avg.}} \\
\cmidrule(lr){3-17}
& &
AUC & AP & EER &
AUC & AP & EER &
AUC & AP & EER &
AUC & AP & EER &
AUC & AP & EER \\
\midrule
$\times$ & $\times$  
& 0.623 & 0.729 & 67.6 
& 0.647 & 0.700 & 40.3 
& 0.576 & 0.595 & 44.7 
& 0.533 & 0.905 & 47.6 
& 0.595 & 0.732 & 50.0 \\

$\checkmark$ & $\times$
& 0.892 & \textbf{0.944} & \textbf{18.6} 
& 0.864 & 0.927 & 22.8 
& 0.829 & 0.861 & 25.2 
& 0.939 & 0.992 & 12.7 
& 0.881 & 0.931 & 19.8 \\

$\times$ & $\checkmark$ 
& 0.617 & 0.742 & 43.0 
& 0.573 & 0.579 & 47.0 
& 0.516 & 0.525 & 49.8 
& 0.511 & 0.894 & 49.0 
& 0.554 & 0.685 & 47.2 \\

$\checkmark$ & $\checkmark$ 
& \textbf{0.895} & 0.943 & 18.9 
& \textbf{0.884} & \textbf{0.939} & \textbf{20.1} 
& \textbf{0.844} & \textbf{0.873} & \textbf{24.1} 
& \textbf{0.948} & \textbf{0.994} & \textbf{11.6} 
& \textbf{0.893} & \textbf{0.937} & \textbf{18.7} \\
\bottomrule
\end{tabular}
\label{Method ablation}
\end{table*}

\noindent\textbf{Effects of the Loss Terms.}
To assess the contribution of each loss component, An ablation study is performed on the subspace orthogonality constraint $\mathcal{L}_{\text{orth}}$ and the spectral consistency constraint $\mathcal{L}_{\text{spec}}$. Specifically, one term is removed at a time while keeping the network architecture and training configuration strictly unchanged to ensure fair comparisons. We evaluate performance using AUC, AP, and EER, with complete results reported in Table~\ref{Loss ablation}. 
The results show that incorporating both $\mathcal{L}_{\text{orth}}$ and $\mathcal{L}_{\text{spec}}$ yields the best performance across all metrics. Compared to the variant without either term, the average AUC increases by approximately 2\%, AP improves by about 1.2\%, and EER decreases by roughly 1.8\%. These gains indicate that the two subspace constraints are synergistic under joint optimization, effectively enhancing discriminative capability and improving training stability.

\begin{table*}[htbp]
\centering
\caption{Ablation study on the effectiveness of $\mathcal{L}_{\text{orth}}$ and $\mathcal{L}_{\text{spec}}$. All models are trained on FF++ (c23)~\cite{Rssler2019FF++} and evaluated in the cross-dataset setting. Metrics are reported as AUC \textbar\ AP \textbar\ EER (frame-level). The average performance across all datasets (Avg.) is additionally reported, and the best results are highlighted in \textbf{bold}.}
\begin{tabular}{c|c|ccc|ccc|ccc|ccc|ccc}
\toprule
\multirow{2}{*}{\raisebox{-0.3em}{$\mathcal{L}_{\text{orth}}$}} &
\multirow{2}{*}{\raisebox{-0.3em}{$\mathcal{L}_{\text{spec}}$}} &
\multicolumn{3}{c|}{\textbf{CDF-v2}} &
\multicolumn{3}{c|}{\textbf{DFDCP}} &
\multicolumn{3}{c|}{\textbf{DFDC}} &
\multicolumn{3}{c|}{\textbf{DFD}} &
\multicolumn{3}{c}{\textbf{Avg.}} \\
\cmidrule(lr){3-17}
& &
AUC & AP & EER &
AUC & AP & EER &
AUC & AP & EER &
AUC & AP & EER &
AUC & AP & EER \\
\midrule
$\times$ & $\times$
& 0.865 & 0.925 & 22.0
& 0.859 & 0.925 & 22.9
& 0.831 & 0.862 & 25.3
& 0.937 & 0.992 & 13.2
& 0.873 & 0.926 & 20.8 \\

$\checkmark$ & $\times$
& 0.869 & 0.929 & 21.7
& 0.868 & 0.928 & 21.7
& 0.823 & 0.856 & 25.5
& 0.945 & 0.993 & 12.4
& 0.876 & 0.927 & 20.3 \\

$\times$ & $\checkmark$
& 0.875 & 0.932 & 20.7
& 0.874 & 0.931 & 20.6
& 0.823 & 0.855 & 26.0
& 0.946 & 0.993 & 12.0
& 0.880 & 0.928 & 19.8 \\

$\checkmark$ & $\checkmark$
& \textbf{0.895} & \textbf{0.943} & \textbf{18.9}
& \textbf{0.884} & \textbf{0.939} & \textbf{20.1}
& \textbf{0.844} & \textbf{0.873} & \textbf{24.1}
& \textbf{0.948} & \textbf{0.993} & \textbf{11.6}
& \textbf{0.893} & \textbf{0.937} & \textbf{18.7} \\
\bottomrule
\end{tabular}
\label{Loss ablation}
\end{table*}

\noindent\textbf{Effects of the Number of Tuned Layers $m$ in FLM.}
Our backbone is CLIP ViT-L/14~\cite{Radford2021CLIP}, which contains 24 Vision Transformer blocks. Each block includes a self-attention module, and each self-attention module consists of four linear projections. Therefore, the backbone contains 96 linear layers that can be adapted.
To study how the number of tuned layers influences performance, we vary $m \in \{1,4,16,48,96\}$ under the same training setup and evaluate frame-level AUC in four cross-dataset scenarios. When $m=96$, this setting is equivalent to enabling parameter updates for all self-attention linear layers.
The results are reported in Table~\ref{m}. As shown in the table, increasing $m$ leads to a clear performance improvement in the early stage, and the best performance is achieved at $m=16$, with an average AUC of \textbf{89.3\%}. This indicates that selecting a moderate number of layers for updating can effectively improve the cross-dataset generalization ability of the model in deepfake detection.
However, when $m>16$, performance slightly degrades, suggesting that excessively enabling parameter updates across layers may increase disturbance to pretrained semantic representations and consequently harm generalization. Based on these observations, $m=16$ is adopted as the default setting in subsequent experiments to achieve a favorable balance between performance and representation preservation.

\begin{table}[htbp]
\centering
\caption{Effect of the number of tuned layers $m$ in FLM on frame-level AUC across datasets and the overall average performance.}
\begin{tabular}{cccccc}
\toprule
\textbf{$m$} & \textbf{CDF-v2} & \textbf{DFDCP} & \textbf{DFDC} & \textbf{DFD} & \textbf{Avg.} \\
\midrule
1 & 0.807 & 0.752 & 0.791 & 0.925 & 0.819 \\
4 & 0.893 & 0.862 & \textbf{0.848} & 0.944 & 0.887 \\
\textbf{16} & \textbf{0.895} & \textbf{0.884} & 0.844 & \textbf{0.948} & \textbf{0.893} \\
48 & 0.884 & 0.875 & 0.828 & 0.939 & 0.881 \\
96 & 0.892 & 0.864 & 0.829 & 0.939 & 0.881 \\
\bottomrule
\end{tabular}
\label{m}
\end{table}

\noindent\textbf{Effects of the Number of Artifact Subspaces $K$.}
This experiment investigates how the number of artifact subspaces $K$ influences performance, revealing the trade-off between subspace granularity and representation capacity. Under the same training setup, we set $K \in \{1,3,5,7,9\}$ and evaluate frame-level AUC in four cross-dataset scenarios.
The results are reported in Table~\ref{K}. As shown in the table, increasing $K$ leads to a clear performance gain in the early stage. However, when $K>5$, the overall performance saturates and slightly degrades. With $K=5$, the model achieves frame-level AUCs of 89.5\%, 88.4\%, 84.4\%, and 94.8\% on CDF-v2, DFDCP, DFDC, and DFD, respectively, yielding the best average performance of \textbf{89.3\%} among all configurations.
These results suggest that a moderate number of artifact subspaces improves the model's ability to capture diverse forgery artifacts, whereas overly fine-grained subspace partitioning may introduce redundant representations and increase optimization difficulty, ultimately harming generalization. Therefore, $K=5$ is adopted as the default setting to achieve a favorable balance between performance and model complexity.

\begin{table}[htbp]
\centering
\caption{Effect of the number of artifact subspaces $K$ on frame-level AUC across datasets and the overall average performance.}
\begin{tabular}{cccccc}
\toprule
\textbf{$K$} & \textbf{CDF-v2} & \textbf{DFDCP} & \textbf{DFDC} & \textbf{DFD} & \textbf{Avg.} \\
\midrule
1 & 0.873 & 0.879 & 0.829 & 0.943 & 0.881 \\
3 & 0.872 & 0.853 & 0.841 & 0.946 & 0.878 \\
\textbf{5} & \textbf{0.895} & \textbf{0.884} & \textbf{0.844} & \textbf{0.948} & \textbf{0.893} \\
7 & 0.888 & 0.874 & 0.832 & 0.946 & 0.885 \\
9 & 0.894 & 0.873 & 0.829 & 0.945 & 0.885 \\
\bottomrule
\end{tabular}
\label{K}
\end{table}

\begin{figure*}[htbp]
    \centering
    \includegraphics[width=0.9\linewidth]{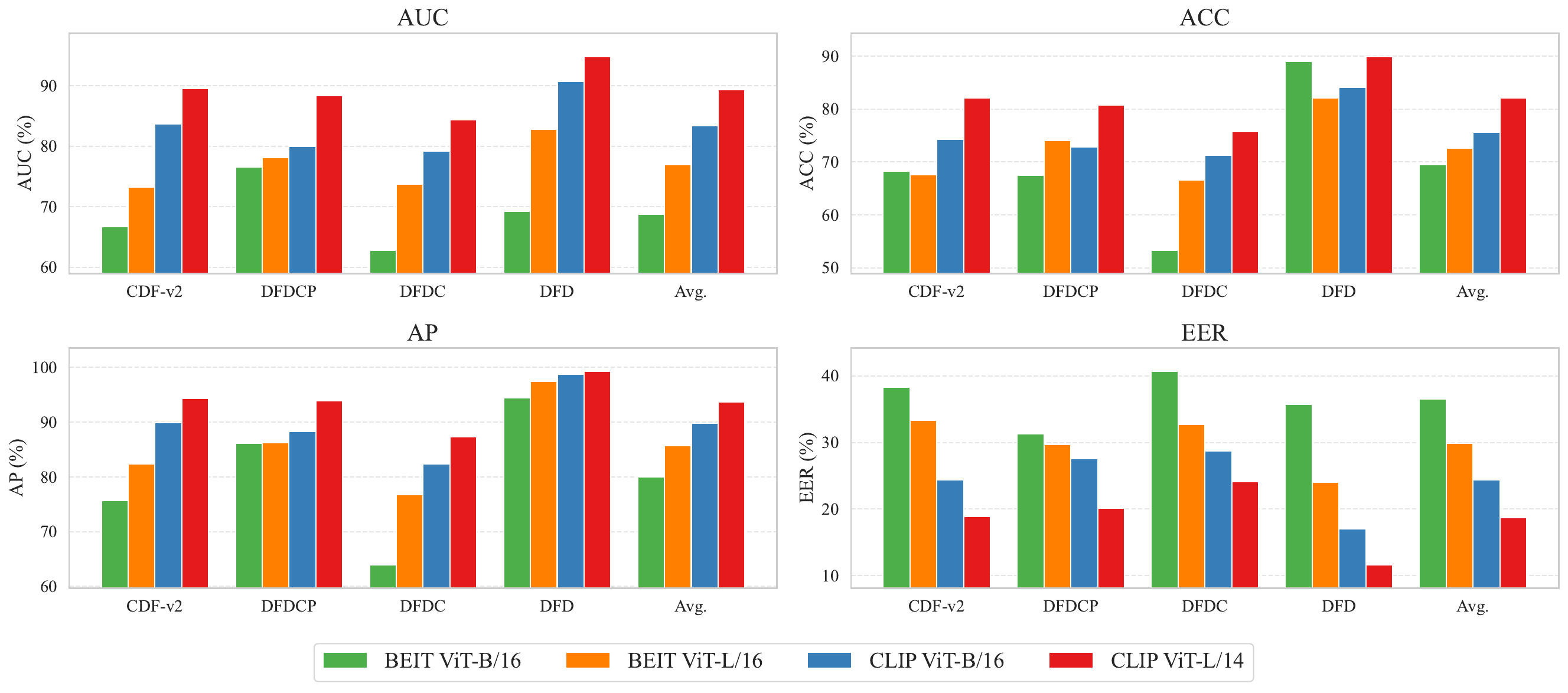}
    \caption{Ablation study on different pretrained vision foundation models (VFMs). All models are trained on FF++ (c23)~\cite{Rssler2019FF++} and evaluated in the cross-dataset setting. Metrics are reported as AUC \textbar \ ACC \textbar \ AP \textbar\ EER (frame-level). The average performance across all datasets (Avg.) is additionally reported.}
    \label{Vfms}
\end{figure*}

\noindent\textbf{Effects of Pretrained Vision Foundation Models.}
This section investigates how different pretrained vision foundation models (VFMs) influence the performance of FMSD. Four widely used VFMs are considered: BEiT ViT-B/16 and BEiT ViT-L/16~\cite{Bao2021BEIT}, as well as CLIP ViT-B/16 and CLIP ViT-L/14~\cite{Radford2021CLIP} (our default configuration). As shown in Fig.~\ref{Vfms}, the CLIP variants consistently outperform the BEiT counterparts across all test datasets in terms of frame-level AUC, ACC, AP, and EER. This gap is primarily attributed to the difference in pretraining paradigms. BEiT learns reconstruction-oriented representations via masked image modeling, emphasizing local pixel-level consistency, whereas CLIP acquires semantically discriminative features through large-scale image--text contrastive pretraining, which better supports semantic transfer for generalizable deepfake detection.
From the perspective of model capacity, ViT-L/14 further improves over ViT-B/16, indicating that a higher-capacity visual encoder better supports multi-artifact subspace decomposition in FMSD for extracting more discriminative forgery representations.

\begin{figure*}[htbp]
    \centering
    % ========= global params =========
    \newcommand{\imgw}{0.14\linewidth}
    \newcommand{\namew}{0.06\linewidth}
    \setlength{\tabcolsep}{2pt}
    \renewcommand{\arraystretch}{1}
    \newcommand{\img}[1]{\raisebox{-0.5\height}{\includegraphics[width=\imgw]{#1}}}
    % ===================== Row 1 =====================
    \begin{minipage}[t]{0.49\linewidth}
        \centering
        \begin{tabular}{m{\namew} c c c c c c}
            & \textbf{Input} & \textbf{S1} & \textbf{S2} & \textbf{S3} & \textbf{S4} & \textbf{S5} \\
            \multirow{2}{*}[-1.5em]{\centering \textbf{DF}}
            & \img{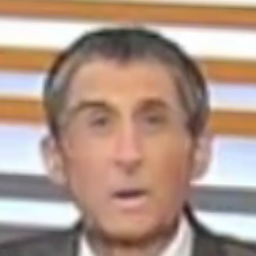}
            & \img{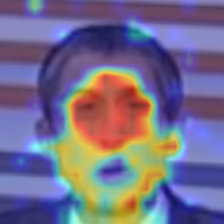}
            & \img{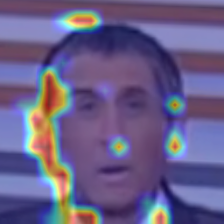}
            & \img{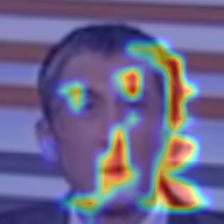}
            & \img{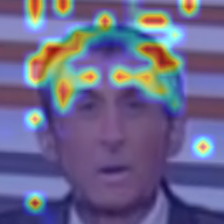}
            & \img{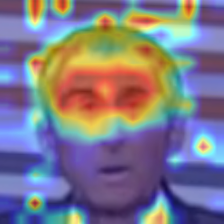} \\
            & \img{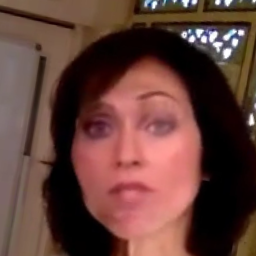}
            & \img{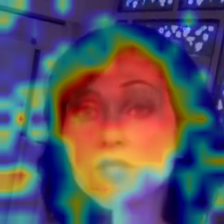}
            & \img{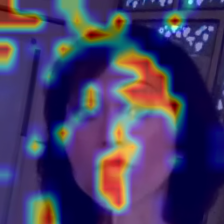}
            & \img{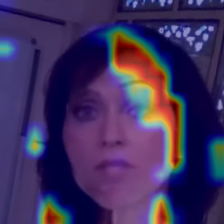}
            & \img{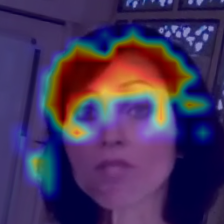}
            & \img{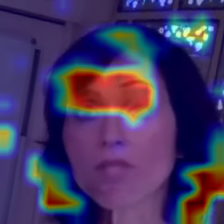}
        \end{tabular}
    \end{minipage}
    \hfill
    \begin{minipage}[t]{0.49\linewidth}
        \centering
        \begin{tabular}{m{\namew} c c c c c c}
            & \textbf{Input} & \textbf{S1} & \textbf{S2} & \textbf{S3} & \textbf{S4} & \textbf{S5} \\
            \multirow{2}{*}[-1.5em]{\centering \textbf{F2F}}
            & \img{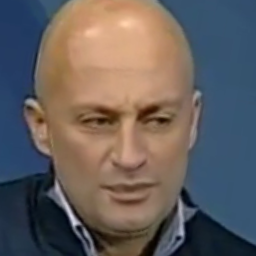}
            & \img{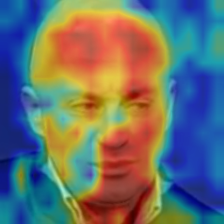}
            & \img{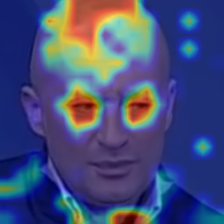}
            & \img{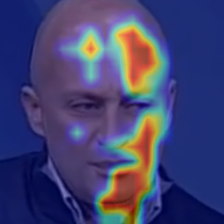}
            & \img{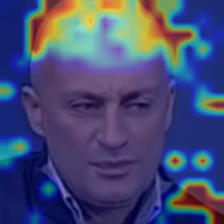}
            & \img{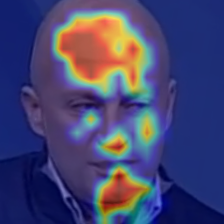} \\
            & \img{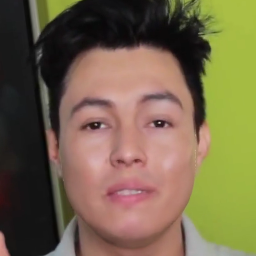}
            & \img{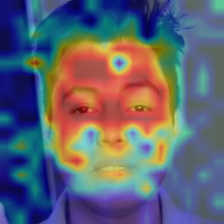}
            & \img{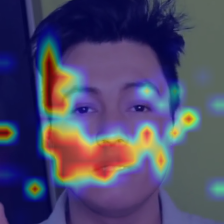}
            & \img{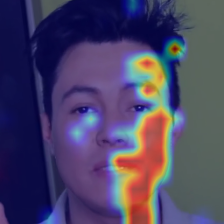}
            & \img{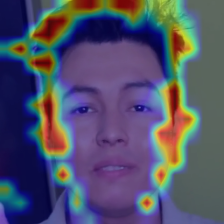}
            & \img{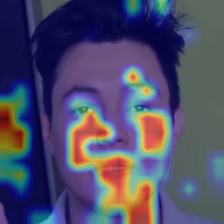}
        \end{tabular}
    \end{minipage}
    \vspace{6pt}
    % ===================== Row 2 =====================
    \begin{minipage}[t]{0.49\linewidth}
        \centering
        \begin{tabular}{m{\namew} c c c c c c}
            \multirow{2}{*}[-1.5em]{\centering \textbf{FS}}
            & \img{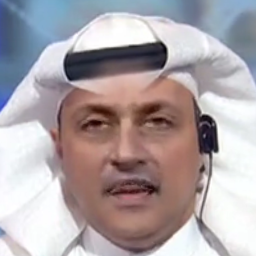}
            & \img{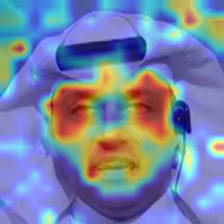}
            & \img{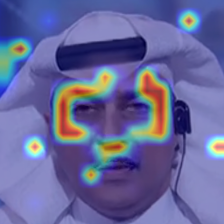}
            & \img{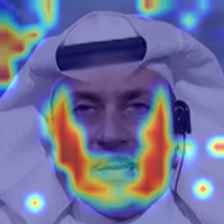}
            & \img{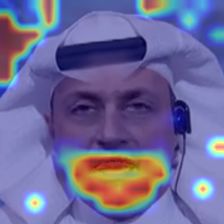}
            & \img{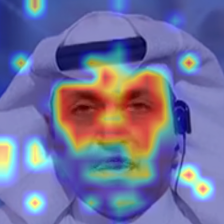} \\
            & \img{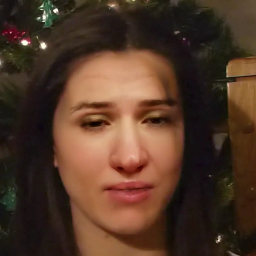}
            & \img{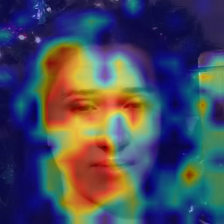}
            & \img{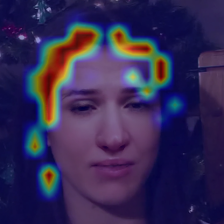}
            & \img{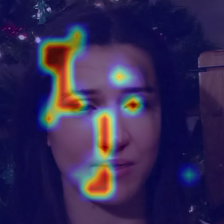}
            & \img{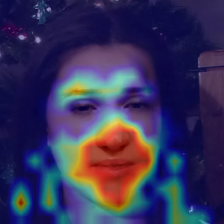}
            & \img{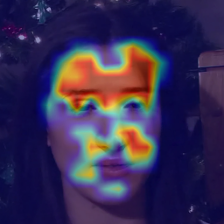}
        \end{tabular}
    \end{minipage}
    \hfill
    \begin{minipage}[t]{0.49\linewidth}
        \centering
        \begin{tabular}{m{\namew} c c c c c c}
            \multirow{2}{*}[-1.5em]{\centering \textbf{NT}}
            & \img{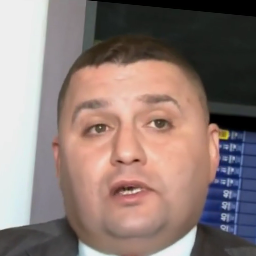}
            & \img{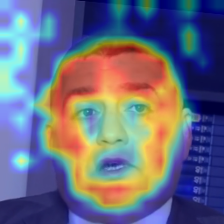}
            & \img{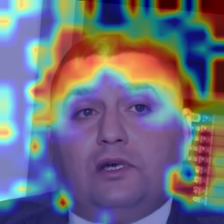}
            & \img{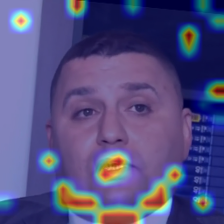}
            & \img{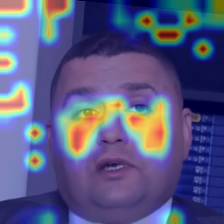}
            & \img{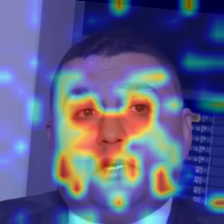} \\
            & \img{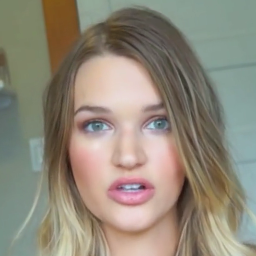}
            & \img{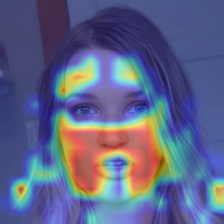}
            & \img{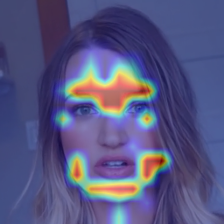}
            & \img{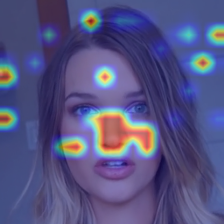}
            & \img{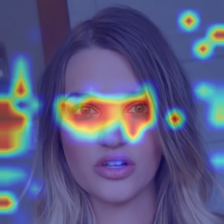}
            & \img{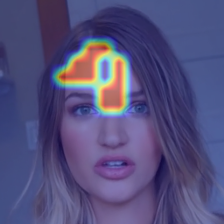}
        \end{tabular}
    \end{minipage}
    \caption{
        Grad-CAM~\cite{Selvaraju2016GradCAMVE} visualizations of different artifact subspaces across the four manipulation subsets in FaceForensics++ (DF, F2F, FS, and NT). Each row corresponds to one subset, and S1--S5 denote different artifact subspaces. The distinct activation patterns indicate that different subspaces capture diverse and complementary forgery cues.
    }
    \label{subspace_comparison}
\end{figure*}

\subsection{Visualization}
To further examine the effect of multi-artifact subspace decomposition, Grad-CAM~\cite{Selvaraju2016GradCAMVE} is used to visualize the forgery-related regions highlighted by different artifact subspaces. For each visualization, only one artifact subspace is retained, while the weights of all remaining subspaces are set to zero, so that the activation pattern of the selected subspace can be examined in isolation. As shown in Fig.~\ref{subspace_comparison}, the artifact subspaces exhibit distinct response distributions across manipulation types. Their activations are concentrated on different facial structures, including blending boundaries and local components such as the eyes, nose, and mouth. In particular, S1 tends to respond to relatively broader facial regions, whereas S2--S5 focus more on different local subregions, including the eyes, nose, mouth, and nearby boundary areas, indicating that different subspaces encode distinct yet complementary forgery cues across manipulation types. These results provide visual evidence that the proposed decomposition strategy effectively separates heterogeneous forgery artifacts and improves the modeling of forgery-related traces.

\section{Conclusion}
In this paper, FMSD, a deepfake detection framework is proposed for improving generalization across datasets and challenging real-world scenarios. The proposed method is motivated by the observation that existing adaptation strategies often overfit forgery-specific cues and undesirably disturb pretrained semantic representations, thereby limiting robustness to unseen manipulations and complex degradations. To address this issue, FMSD combines forgery-aware layer masking with multi-artifact subspace decomposition. Specifically, forgery-aware layer masking generates layer-wise masks based on the bias-variance characteristics of layer-wise gradients to control whether each layer participates in parameter updates, thereby reducing unnecessary disturbance to pretrained representations. Building upon this, multi-artifact subspace decomposition decomposes the selected layer weights into one semantic subspace and multiple learnable artifact subspaces, enabling structured modeling of heterogeneous forgery artifacts. In addition, orthogonality and spectral consistency constraints are introduced to regularize the learned subspaces and preserve the spectral structure of pretrained weights.

Extensive experiments under multiple evaluation protocols demonstrate that FMSD consistently outperforms existing methods in cross-dataset evaluation, cross-manipulation detection, real-world scenarios, and robustness testing under diverse distortions. These findings suggest that explicitly separating semantic representations from forgery-related artifact representations is an effective way to improve the generalization capability of deepfake detectors. We hope this work can provide a useful perspective for developing more robust deepfake detection methods in open and complex environments.

\bibliographystyle{IEEEtran}
\bibliography{Reference}

\end{document}